\def\year{2020}\relax
\documentclass[letterpaper]{article} 
\usepackage{aaai20}  
\usepackage{times}  
\usepackage{helvet} 
\usepackage{courier}  
\usepackage[hyphens]{url}  
\usepackage{graphicx} 
\usepackage{graphicx}  
\usepackage[whole]{bxcjkjatype}
\usepackage{amssymb}
\usepackage{algorithmic,algorithm}
\usepackage{amsmath}
\usepackage{url}
\newcommand{\1}{\mbox{1}\hspace{-0.25em}\mbox{l}}
\newcommand{\argmax}{\mathop{\mathrm{arg~max}}\limits}

\title{Domain Generalization Using a Mixture of Multiple Latent Domains}
\author{Toshihiko Matsuura\textsuperscript{\rm 1}, Tatsuya Harada\textsuperscript{\rm 1}\textsuperscript{,}\textsuperscript{\rm 2}\\
\textsuperscript{\rm 1}The University of Tokyo
\textsuperscript{\rm 2}RIKEN\\
\{matsuura, harada\}@mi.t.u-tokyo.ac.jp
}
\begin{document}

\maketitle

\begin{abstract}
When domains, which represent underlying data distributions, vary during training and testing processes, deep neural networks suffer a drop in their performance. Domain generalization allows improvements in the generalization performance for unseen target domains by using multiple source domains. Conventional methods assume that the domain to which each sample belongs is known in training. However, many datasets, such as those collected via web crawling, contain a mixture of multiple latent domains, in which the domain of each sample is unknown. This paper introduces domain generalization using a mixture of multiple latent domains as a novel and more realistic scenario, where we try to train a domain-generalized model without using domain labels. To address this scenario, we propose a method that iteratively divides samples into latent domains via clustering, and which trains the domain-invariant feature extractor shared among the divided latent domains via adversarial learning. We assume that the latent domain of images is reflected in their style, and thus, utilize style features for clustering. By using these features, our proposed method successfully discovers latent domains and achieves domain generalization even if the domain labels are not given. Experiments show that our proposed method can train a domain-generalized model without using domain labels. Moreover, it outperforms conventional domain generalization methods, including those that utilize domain labels.
\end{abstract}

\begin{figure}[t]
\centering
\includegraphics[width=0.98\columnwidth]{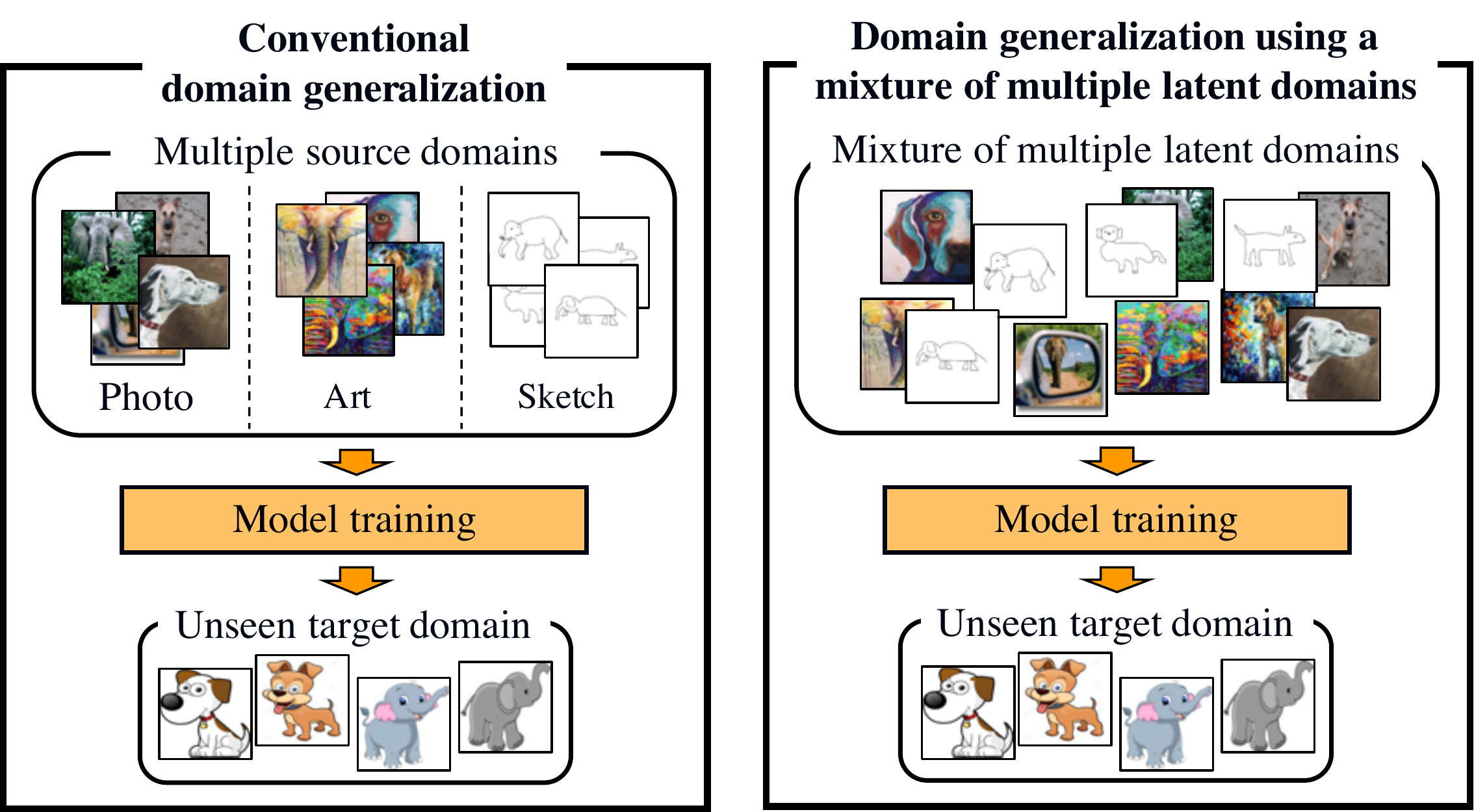}
\caption{Unlike conventional domain generalization, domain generalization using a mixture of multiple latent domains aims to train a domain-generalized model without domain labels (e.g., Photo, Art, Sketch), which represent the domain to which each sample belongs.}
\label{fig:data}
\end{figure}

\section{Introduction}
In the development of deep neural networks (DNNs), many methods that achieve good performance in computer vision tasks have been proposed \cite{FasterRCNN,deeplab_v3+}.
A domain represents an underlying data distribution, and these methods assume that the domains given in training (\textit{source domain}) and in testing (\textit{target domain}) are the same. However, it is known that DNNs suffer a drop in their performance due to domain shift \cite{dataset_bias}.\par
To address this problem, extensive research has been carried out on domain generalization, which aims to train a domain-generalized model that performs well for the unseen target domain by using labeled data from multiple source domains. Considering the situation where a DNN is used for autonomous driving or robots in the real world, it is desirable to perform well under different conditions (e.g., illumination, types of objects) from the data given in training. 
Because we can access no samples in the target domain, domain generalization can be considered a more difficult and a more important task than domain adaptation \cite{DAN,GRL}, in which we can access labeled/unlabeled samples of the target domain in training.\par
To achieve domain generalization, several domain generalization methods have been proposed, including methods that train the feature extractor so that the feature distributions among multiple source domains are matched \cite{MMD_AAE,CIDDG}, or methods that train models for each domain and which combine them in testing \cite{SSN,D_SAM}.
These conventional methods require domain labels, which represent the domain to which each sample in multiple source domains belongs.
However, most datasets, such as those collected via web crawling, are a mixture of multiple latent domains, and it is difficult to know the domain labels. For example, there are several types of image search results for ``dog'', such as close-up photos of a face, photos of a dog figure in nature, and drawings of a dog. In this scenario, domain labels have to be attached manually to use conventional methods, but this process may be costly and time-consuming. Moreover, it is not obvious how to divide a mixture of multiple latent domains into each domain because those underlying data distributions are unknown.\par
JiGen \cite{JiGen} achieves domain generalization without domain labels by combining supervised learning and self-supervised learning to solve jigsaw puzzles of the training images. 
However, it does not take advantage of the fact that there exist several latent domains in the source domain. Therefore, in this paper, we propose a novel and realistic scenario called \textit{domain generalization using a mixture of multiple latent domains}, in which the source domain contains multiple latent domains, and the domain to which each sample belongs is unknown. As shown in Fig.~\ref{fig:data}, in the proposed scenario, we try to train a model that performs well for the unseen target domain using a mixture of multiple latent domains. Moreover, we propose a novel method to solve this scenario. First, we assume that the latent domain of images is reflected in their style. Although other factors may also be considered, such as the background, location, and pose change, domain mismatches may be more severe when image styles are different, such as photos, NIR images, paintings, or sketches. Therefore, we utilize style features proposed in the research field of style transfer as domain-discriminative features to discover latent domains. Specifically, we utilize a stack of convolutional feature statistics (i.e., mean and standard deviation) that are known to be capable of capturing image styles \cite{Li}.
Once domain-discriminative features are obtained, our method iteratively assigns pseudo domain labels by clustering them, and trains a domain-invariant feature extractor shared among multiple latent domains by adversarial learning.\par
Experiments with benchmark datasets show that our proposed method is effective for domain generalization using a mixture of multiple latent domains, and it outperforms conventional domain generalization methods that use domain labels. Moreover, it is found that the use of pseudo domain labels obtained by clustering style features improves the classification performance compared with the use of original domain labels annotated by humans.\par

\section{Related Work}
Here, we explain domain adaptation and domain generalization methods.
Moreover, we explain style-transfer methods because as domain-discriminative features, our proposed method utilizes style features that were originally proposed in the research field of style transfer.
\subsection{Domain Adaptation}
To deal with domain shift \cite{dataset_bias}, domain adaptation and domain generalization have been studied. Domain adaptation aims to generalize a model from the source domain to the target domain with data in both domains. In unsupervised domain adaptation, several methods are employed to match the distribution in pixel space \cite{PixDA,CrDoCo} or feature space \cite{DAN,GRL}.
Although these methods assume single-source and target domains, multi-source domain adaptation methods \cite{DCTN,MultiDA} utilize multiple source domains for domain adaptation to learn domain relations.\par
Moreover, for the case in which the domains to which each sample belongs are unknown, Mancini et al. \cite{mDA_layer} proposed a deep architecture that automatically discovers multiple latent domains, and it uses this information to align the distributions of the internal feature representations of sources and target domains. In contrast to our proposed method, this method is suitable for domain adaptation, and requires target samples in training.\par
\subsection{Domain Generalization}
Domain generalization aims to train a domain-generalized model for the unseen target domain by using multiple source domains. Unlike domain adaptation, target samples are not given in training. The representative methods for domain generalization match the feature distributions among multiple source domains by using an auto-encoder \cite{D_MTAE,MMD_AAE} or using adversarial learning \cite{CIDDG,PAD}. In addition, several methods have been proposed, such as a method that is based on meta learning \cite{MLDG,MetaReg}, one that uses domain-specific aggregation modules \cite{D_SAM}, and a method that combines supervised learning and self-supervised learning to solve jigsaw puzzles \cite{JiGen}.\par
Most conventional domain generalization methods require domain labels, which represent the domains to which each sample belongs. However, in the scenario of domain generalization using a mixture of multiple domains, we cannot apply these methods because domain labels are not given. Although JiGen \cite{JiGen} does not require domain labels in training, it is different from our proposed method, which assumes that the source domain contains multiple latent domains and take advantage of them.

\begin{figure*}[t]
\centering
\includegraphics[width=2.08\columnwidth]{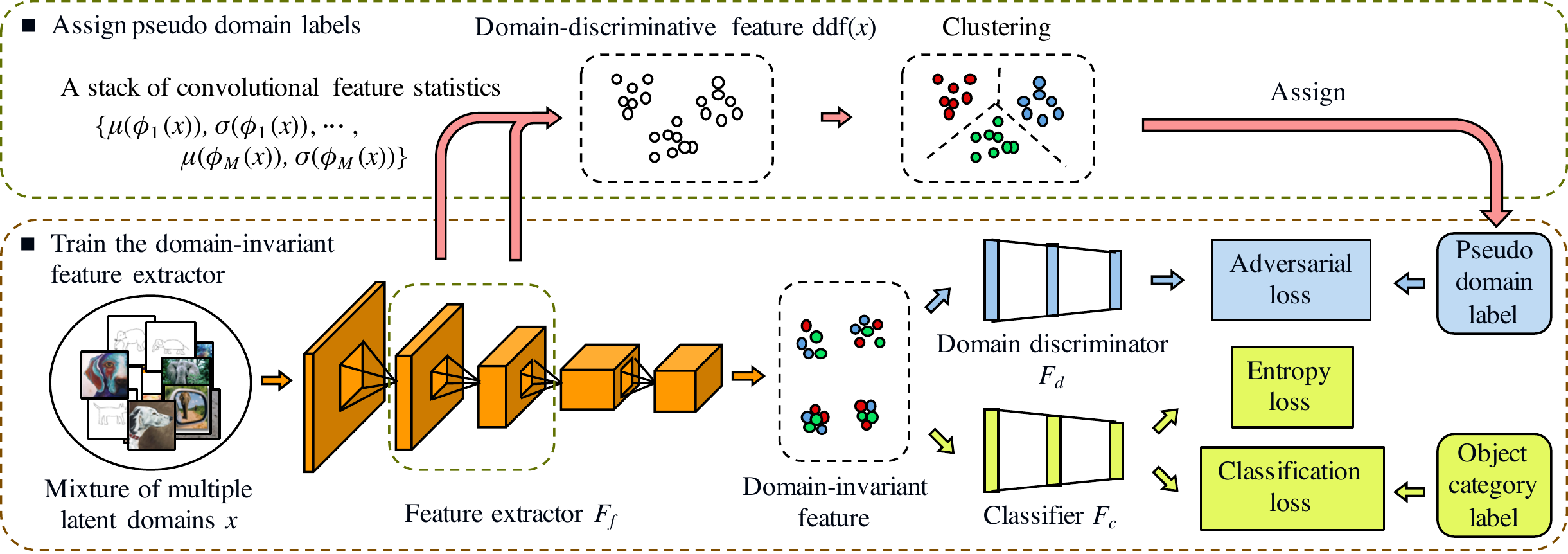}
\caption{Illustration of our proposed method: Our method iteratively assigns pseudo domain labels by clustering domain-discriminative features extracted from lower layers of the feature extractor, and trains the domain-invariant feature extractor via adversarial learning.}
\label{fig:model}
\end{figure*}

\subsection{Style Transfer}
Style transfer enables us to transfer the style of an image called \textit{style image} to that of an image called \textit{content image} while preserving its content. Neural style transfer \cite{Gatys} utilizes Gram matrices of the neural activations from different layers of a convolutional neural network (CNN) to represent the artistic style of an image. Li et al. \cite{Li} theoretically showed that matching the Gram matrices of neural activations is equivalent to minimizing the maximum mean discrepancy with the second-order polynomial kernel, and constructed another style loss by aligning the convolutional feature statistics (i.e., mean and standard deviation) of two feature maps between style and generated images.
AdaIN \cite{AdaIN} enables arbitrary style transfer in real-time by replacing the convolutional feature statistics of the content image with those of the style image.
Inspired by these methods, we assume that the latent domain of images is reflected in their style and utilize convolutional feature statistics as domain-discriminative features.

\section{Domain Generalization Using a Mixture of Multiple Latent Domains}
In conventional domain generalization, the model trained with $K$ source domains $\mathcal{D}_s=\{\mathcal{D}_s^k\}_{k=1}^K$, which share the same tasks (input $x$ and label spaces $y$) but have different data distributions, accurately works for the new target domain $\mathcal{D}_t$. In this paper, we focus on the image classification task and set the number of object categories to $C$. Moreover, when the $k$-th source domain $\mathcal{D}_s^k$ has $N_s^k$ samples, the dataset given in training is $\mathcal{D}_s=\{\mathcal{D}_s^k\}_{k=1}^K, \mathcal{D}_s^k=\{(x_i^k,y_i^k)\}_{i=1}^{N_s^k}$.
This can also be represented using $\mathcal{D}_s=\{(x_i,y_i,d_i)\}_{i=1}^{N_s}$, when the domain to which each sample belongs and the total number of samples included in all source domains are defined as $d_i$ and $N_s$, respectively. Namely, conventional domain generalization methods train the model that works well for the unseen target domain by using input images $x_i$, object category labels $y_i$, and domain labels $d_i$.\par
However, as we described above, a real dataset may be a mixture of multiple latent domains, and it is difficult to obtain domain labels in this case. Therefore, we propose a scenario called domain generalization using a mixture of multiple latent domains, where the given dataset is $\mathcal{D}_s=\{(x_i,y_i)\}_{i=1}^{N_s}$ because domain labels $d_i$ are unknown.\par

\section{Proposed Method}
In this section, we explain the details of our proposed method. An overview of our method is shown in Fig.~\ref{fig:model}. Our method utilizes adversarial learning with a domain discriminator to train the domain-invariant feature extractor from among multiple latent domains; this approach is also used in conventional domain adaptation or generalization methods \cite{GRL,CIDDG}. Although adversarial domain generalization methods require domain labels, they are not given in domain generalization using a mixture of multiple latent domains. Therefore, our method iteratively reassigns pseudo domain labels by clustering domain-discriminative features obtained from the model.\par
The key point is how to extract domain-discriminative features from the model in order to cluster samples by their latent domains. Clustering features obtained from the model may generally divide samples by their object categories, and not by their domains. Moreover, our method aims to train a domain-invariant feature extractor by making the outputs domain-invariant, which hinders the extraction of domain-discriminative features from the model. To solve this problem, we assume that the latent domain of images is reflected in their style, and we thus propose to utilize style features used in style transfer. Specifically, we utilize a stack of convolutional feature statistics (i.e., mean and standard deviations) obtained from lower layers of the feature extractor. In this way, our method can divide samples into each latent domain and achieve domain generalization. In the rest of the section, we describe the details of each component of our proposed method.
\subsection{Adversarial Domain Generalization}
Adversarial learning, which is developed from generative adversarial networks (GANs) \cite{GAN}, has been used for research in domain adaptation \cite{GRL} and generalization \cite{CIDDG}.
Generally, a deep learning model can be divided into a feature extractor $F_f$ and a classifier $F_c$. These models can be trained with the following classification loss $L_{\mathrm{cls}}$.
\begin{equation}
   L_{\mathrm{cls}}=-\frac{1}{N_s}\sum_{i=1}^{N_s}\sum_{c=1}^{C}\1_{[c=y_i]}\log F_c(F_f(x_i))
    \label{eq:class}
\end{equation}
\par
In addition to these components, adversarial learning introduces a domain discriminator $F_d$, which is trained to discriminate the domains when outputs of the feature extractor are inputted.
Conversely, the feature extractor is trained to extract features that make it difficult for the domain discriminator to discriminate their domains.
This makes it possible to extract domain-invariant features from among multiple source domains, which generalizes the model for the unseen target domain.
The adversarial loss $L_{\mathrm{adv}}$ is defined as follows.
\begin{equation}
    L_{\mathrm{adv}}=-\frac{1}{N_s}\sum_{i=1}^{N_s}\sum_{k=1}^{\hat{K}}\1_{[k=\hat{d}_i]}\log F_d(F_f(x_i))
    \label{eq:adv}
\end{equation}
Although conventional methods use known domain labels $d_i$ and the known number of domains $K$, our proposed method uses pseudo domain labels $\hat{d}_i$ by assigning samples into $\hat{K}$ pseudo domains using clustering.\par
It is known that adversarial learning tends to generate ambiguous features near the decision boundary by trying to simply match the distributions among multiple source domains \cite{MCD}. Therefore, we introduce the entropy loss $L_{\mathrm{ent}}$ \cite{entropy}, which is used in some domain adaptation methods \cite{RTN,SymNets} to train a more discriminative model for target samples by encouraging low-density separation between object categories. Although previous domain adaptation methods adapt it to only unlabeled target samples, our method adapts it to all labeled source samples as follows.
\begin{equation}
L_{\mathrm{ent}}=-\frac{1}{N_s}\sum_{i=1}^{N_s}H(F_c(F_f(x_i)))
\label{eq:entropy}
\end{equation}
Here, $H(\cdot)$ represents the entropy function. This entropy loss enables us to extract discriminative features for object categories and to improve the classification accuracy.\par
The total training objective is described as follows.
\begin{align}
    \min_{F_f,F_c}&=L_{\mathrm{cls}}(F_f,F_c)+\lambda(L_{\mathrm{ent}}(F_f,F_c)-L_{\mathrm{adv}}(F_f,F_d))\nonumber\\
    \min_{F_d}&=L_{\mathrm{adv}}(F_f,F_d)
    \label{eq:total}
\end{align}
Here, $\lambda$ denotes the trade-off parameter to suppress the noise signal of two losses $L_{\mathrm{adv}}, L_{\mathrm{ent}}$ in the early stage of training.

\subsection{Domain-discriminative Features}
As domain-discriminative features, we utilize style features proposed in the style transfer~\cite{Gatys,Li,AdaIN}. Style transfer aims to generate a stylized image given a content image and a reference style image.
Li et al.~\cite{Li} proposed a new style loss $L_\mathrm{sty}$ to align the convolutional feature statistics (i.e., mean and standard deviation) between the generated image $x_\mathrm{gen}$ and the style image $x_{\mathrm{sty}}$ as follows.

\begin{equation}
\begin{split}
L_\mathrm{sty}=\sum_{m=1}^M\|\mu(&\phi_m(x_\mathrm{gen}))-\mu(\phi_m(x_\mathrm{sty}))\|_2+\\
&\sum_{m=1}^M\|\sigma(\phi_m(x_\mathrm{gen}))-\sigma(\phi_m(x_\mathrm{sty}))\|_2
\end{split}
\end{equation}
Here, each $\phi_m(x)$ denotes the output in a layer used to compute the style loss, and mean $\mu(x)$ and standard deviation $\sigma(x)$ are calculated across spatial dimensions independently for each channel $c$.
\begin{align}
\mu_{c}(x)&=\frac{1}{HW}\sum_{h=1}^{H}\sum_{w=1}^Wx_{chw}\\
\sigma_{c}(x)&=\sqrt{\frac{1}{HW}\sum_{h=1}^{H}\sum_{w=1}^{W}(x_{chw}-\mu_{c}(x))^2+\epsilon}
\end{align}\par
In our method, we assume that the latent domain of images is reflected in their style, and we thus utilize convolutional feature statistics as domain-discriminative features. Further, to combine multi-scale style features obtained from different convolutional layers, we define a stack of them as domain-discriminative features.
Namely, the domain-discriminative feature $\mathrm{ddf}(x)$ is calculated using multiple layers' outputs $\phi_1(x),\cdots,\phi_M(x)$ as follows.
\begin{equation}
\scalebox{0.98}{$
\mathrm{ddf}(x)=\{\mu(\phi_1(x)),\sigma(\phi_1(x)),\cdots,\mu(\phi_M(x)),\sigma(\phi_M(x))\}
$}
\label{eq:feature}
\end{equation}

\begin{algorithm}[t]
\caption{Training algorithm.}
\label{alg:procedure}
\begin{algorithmic}
\REQUIRE{Data: $\mathcal D_s=\left\{\left(x_i^s,y_i^s\right)\right\}_{i=1}^{N_s}$}
\STATE{Initialize $\hat{d}_i,\hat{d}_i^\prime$ with zero}
\WHILE{not end of epoch}
\STATE{Calculate $\{\mathrm{ddf}(x_i)\}_{i=1}^{N_s}$ using Eq.~\ref{eq:feature}}
\STATE{Obtain $\{a_i\}_{i=1}^{N_s}$ by clustering $\{\mathrm{ddf}(x_i)\}_{i=1}^{N_s}$}
\STATE{Calculate $\hat{\pi}$ using Eq.~\ref{eq:perm}} 
\STATE{Update $\hat{d}_i$ with $\hat{\pi}(a_i)$}
\WHILE{not end of minibatch}
\STATE{Sample a minibatch of $x_i, y_i, \hat{d}_i$}
\STATE{Update parameters using Eq.~\ref{eq:total}}
\ENDWHILE
\STATE{Update $\hat{d}_i^{\prime}$ with $\hat{d}_i$}
\ENDWHILE
\end{algorithmic}
\end{algorithm}

\subsection{Training Procedure}
After obtaining domain-discriminative features for all training samples using Eq.~\ref{eq:feature}, our method divides them into $\hat{K}$ clusters by clustering, and utilizes the cluster assignments $a_i$ as pseudo domain labels $\hat{d}_i$. We use a standard clustering algorithm, k-means \cite{Kmeans}, although other clustering algorithms can be used in our method. The overall training procedure is shown in Alg.~\ref{alg:procedure}. Our method iteratively reassigns pseudo domain labels in training.
This is because domain-discriminative features can be extracted more successfully as the training progresses.
In particular, we determine that the reassignment of pseudo domain labels is conducted for each epoch.\par
The problem here is that clustering can divide samples into each cluster but cannot properly decide which domain label should be assigned to each cluster.
If the reassigned pseudo domain labels are shifted largely with those before one epoch, it negatively impacts the training. Therefore, we use the following equation to convert the cluster assignment $a_i$ into the pseudo domain label $\hat{d}_i$ by calculating the permutation $\hat{\pi}$ so as to maximize the rate of agreement between the cluster assignments $\{a_i\}_{i=1}^{N_s}$ and pseudo domain labels before one epoch $\{\hat{d}_i^{\prime}\}_{i=1}^{N_s}$.
\begin{equation}
\hat{\pi}=\argmax_{\pi\in\Pi}\frac{1}{N_s}\sum_{i=1}^{N_s}\1_{[\hat{d}_i^{\prime}=\pi(a_i)]}
\label{eq:perm}
\end{equation}
Here, the optimal permutation $\hat{\pi}$ can be computed using the Kuhn-Munkres algorithm \cite{Kuhn_Munkres}.

\section{Experiments}
\subsection{Datasets}
To evaluate our proposed method, we perform experiments using two datasets for domain generalization.\footnote[1]{The code is publicly available at \url{https://github.com/mil-tokyo/dg_mmld/}.} PACS \cite{PACS} consists of four domains (i.e., Photo, Art Paintings, Cartoon, and Sketch), spanning different image styles, with seven object categories.
VLCS \cite{dataset_bias} aggregates images of five shared object categories (bird, car, chair, dog, and person) from PASCAL VOC 2007 \cite{VOC2007}, LabelMe \cite{labelme}, Caltech-101 \cite{caltech_101}, Sun09 datasets \cite{sun09} which are considered as four separate domains.
Unlike PACS, VLCS provides only photo images with different camera types or composition bias. By using the VLCS dataset, we verify whether our method, which focuses on the image styles, can also deal with domain shifts inside photos.\par
Following the previous work \cite{JiGen}, we use three domains as the source domain, and the other as the target. For the same reason, we split 10\% (in the case of PACS) and 30\% (in the case of VLCS) of the source samples as validation datasets.
In testing, all target samples are used to calculate the accuracy of the model that achieves the best accuracy in the validation dataset. Because domain labels are not given in domain generalization using a mixture of multiple latent domains, we do not use them when using our method.
\subsection{Implementation Details}
As the feature extractor, we use AlexNet and ResNet-18 pre-trained on ImageNet by removing the last layer. As the classifier, we initialize one fully connected layer to have the same number of inputs as before, and to have the same number of outputs as the number of object categories.
As the domain discriminator, we use three fully connected layers ($1024\mathalpha{\rightarrow}1024\mathalpha{\rightarrow}\hat{K}$).
Note that we weight the loss function in Eq.~\ref{eq:adv} by the inverse of the size of pseudo domain labels.
This is because if the number of images per pseudo domain is highly imbalanced, minimizing Eq.~\ref{eq:adv} results in a trivial parametrization where the model will predict the same output regardless of the input.
To acquire the domain-discriminative features of Eq.~\ref{eq:feature}, we use \texttt{relu2} and \texttt{relu3} in the case of AlexNet, and \texttt{conv2\_x} and \texttt{conv3\_x} in the case of ResNet-18.
To conduct adversarial learning in Eq.~\ref{eq:total}, we insert a gradient reversal layer (GRL) \cite{GRL} between the feature extractor and the domain discriminator, and we use the same schedule for $\lambda$ of Eq.~\ref{eq:total} as follows: $\lambda=\frac{2}{1+\exp(-10\cdot p)}-1$.
Here, $p$ is linearly changed from $0$ to $1$ as training progresses.
To reduce the computational cost of clustering, we reduce the dimension of domain-discriminative features to $256$.\par
Basically, we utilize the other hyper-parameters employed by JiGen \cite{JiGen}. In other words, we train the model for $30$ epochs using the mini-batch stochastic gradient descent (SGD) with a momentum of $0.9$, a weight decay of $5e-4$, and a batch size of $128$.
We set the initial learning rate to $1e-3$, and scale it by a factor of $0.1$ after $80\%$ of the training epochs.
In the experiment with the VLCS dataset, we set the initial learning rate to $1e-4$ because it is observed that a high learning rate causes early convergence and over-fitting in the source domain.
Moreover, we set the learning rate of the classifier and the domain discriminator to be 10 times larger than that of the feature extractor because they are trained from scratch.
For pre-processing, we crop images to random sizes and aspect ratios, horizontally flip them randomly, change their brightness/contrast/saturation/hue randomly, and normalize them using ImageNet's statistics.

\subsection{Baselines}
We compare our method with the following recent domain generalization methods.
Deep All: Pre-trained Alexnet or ResNet-18 fine-tuned on the aggregation of all source domains with only the classification loss.
TF \cite{PACS}: The low-rank parameterized neural network, which reduces the number of parameters to be trained.
CIDDG \cite{CIDDG}: The conditional-invariant deep domain generalization method, which matches conditional distributions by considering the changes in the class prior.
MLDG \cite{MLDG}: The meta-learning method by meta-optimization on simulated train/test splits with the domain shift.
CCSA \cite{CCSA}: The deep model in mixture with the classification and contrastive semantic alignment loss to address supervised domain adaptation and generalization.
MMD-AAE \cite{MMD_AAE}: A model that trains feature representations by jointly optimizing a multi-domain autoencoder regularized by the maximum mean discrepancy distance, a discriminator, and a classifier with adversarial learning.
SLRC \cite{SLRC}: The structured low-rank constraint to transfer the knowledge between domain-specific networks and the domain-invariant one.
D-SAM \cite{D_SAM}: Domain-specific aggregation modules, which enable us to merge generic and specific information in an effective manner using an aggregation layer strategy.
JiGen \cite{JiGen}: Jigsaw puzzle-based generalization method, which focuses on the unsupervised task to solve jigsaw puzzles.\par
Note that methods other than Deep All and JiGen cannot be applied for domain generalization using a mixture of multiple latent domains because they require domain labels in training.
Therefore, for these methods, we use the score in the scenario of general domain generalization where domain labels are given.

\begin{table}[t]
    \centering
    \resizebox{\columnwidth}{!}{
    \begin{tabular}{ccccc|c} \hline
     PACS & Art. & Cartoon & Sketch & Photo & Avg.\\ \hline
     \multicolumn{6}{c}{AlexNet}\\ \hline
     Deep All & 63.30 & 63.13 & 54.07 & 87.70 & 67.05 \\
     TF$^\ast$ & 62.86 & 66.97 & 57.51 & \textbf{89.50} & 69.21 \\ \hline
     Deep All & 57.55 & 67.04 & 58.52 & 77.98 & 65.27 \\
     CIDDG$^\ast$ & 62.70 & 69.73 & 64.45 & 78.65 & 68.88 \\ \hline
     Deep All & 64.91 & 64.28 & 53.08 & 86.67 & 67.24 \\
     MLDG$^\ast$ & 66.23 & 66.88 & 58.96 & 88.00 & 70.01 \\ \hline
     Deep All & 64.44 & 72.07 & 58.07 & 87.50 & 70.52 \\
     D-SAM$^\ast$ & 63.87 & 70.70 & 64.66 & 85.55 & 71.20 \\ \hline
     Deep All & 66.68 & 69.41 & 60.02 & \underline{89.98} & 71.52 \\
     JiGen & 67.63 & 71.71 & 65.18 & 89.00 & 73.38 \\ \hline
     Deep All & 68.09 & 70.23 & 61.80 & 88.86 & 72.25 \\ 
     Ours ($\hat{K}\mathalpha{=}2$) & 66.99 & 70.64 & \textbf{67.78} & 89.35 & 73.69 \\ 
     Ours ($\hat{K}\mathalpha{=}3$) & \textbf{69.27} & \textbf{72.83} & 66.44 & 88.98 & \textbf{74.38}\\
     Ours ($\hat{K}\mathalpha{=}4$) & 68.84 & 72.53 & 65.90 & 88.75& 74.01\\ \hline
     
    \multicolumn{6}{c}{ResNet-18}\\ \hline
    Deep All & 77.87 & 75.89 & 69.27 & 95.19 & 79.55 \\ 
    D-SAM$^\ast$ & 77.33 & 72.43 & \textbf{77.83} & 95.30 & 80.72 \\ \hline
    Deep All & 77.85 & 74.86 & 67.74 & 95.73 & 79.05 \\
    JiGen & 79.42 & 75.25 & 71.35 & 96.03 & 80.51 \\ \hline
    Deep All & 78.34 & 75.02 & 65.24 & \underline{96.21} & 78.70 \\
    Ours ($\hat{K}\mathalpha{=}2$) & \textbf{81.28} & \textbf{77.16} & 72.29 & \textbf{96.09} & \textbf{81.83}\\ 
    Ours ($\hat{K}\mathalpha{=}3$) & 79.64 & 76.75 & 71.22 & 95.86 & 80.87\\
    Ours ($\hat{K}\mathalpha{=}4$) & 80.07 & 75.06 & 74.21 & 95.73 & 81.26 \\ \hline
    \end{tabular}
    }
    \caption{Results in the PACS dataset. The title of each column indicates the name of the domain used as the target. The methods with an asterisk use domain labels, but Deep All, JiGen, and our method do not use them. The respective scores are obtained from each method's original paper.}
    \label{tab:PACS}
\end{table}

\begin{table}[t]
    \centering
    \resizebox{\columnwidth}{!}{
    \begin{tabular}{ccccc|c} \hline
    VLCS & Caltech & Labelme & Pascal & Sun & Avg.\\ \hline
     \multicolumn{6}{c}{AlexNet}\\ \hline
     Deep All & 85.73 & 61.28 & 62.71 & 59.33 & 67.26 \\
     CIDDG$^\ast$ & 88.83 & 63.06 & 64.38 & 62.10 & 69.59 \\ \hline
     Deep All & 86.10 & 55.60 & 59.10 & 54.60 & 63.85 \\
     CCSA$^\ast$ & 92.30 & 62.10 & 67.10 & 59.10 & 70.15 \\ \hline
     Deep All & 86.67 & 58.20 & 59.10 & 57.86 & 65.46 \\
     SLRC$^\ast$ & 92.76 & 62.34 & 65.25 & 63.54 & 70.97 \\ \hline
     Deep All & 93.40 & 62.11 & 68.41 & 64.16 & 72.02 \\
     TF$^\ast$ & 93.63 & \textbf{63.49} & 69.99 & 61.32 & 72.11 \\ \hline
     MMD-AAE$^\ast$ & 94.40 & 62.60 & 67.70 & 64.40 & 72.28 \\ \hline
     Deep All & 94.45 & 57.45 & 66.06 & 65.87 & 71.08 \\
     D-SAM$^\ast$ & 91.75 & 56.95 & 58.59 & 60.84 & 67.03 \\ \hline
     Deep All & 96.63 & 59.18 & 71.96 & 62.57 & 72.66 \\
     JiGen & 96.93 & 60.90 & 70.62 & 64.30 & 73.19 \\ \hline
     Deep All & 95.89 & 57.88 & 72.01 & 67.76 & 73.39 \\ 
     Ours ($\hat{K}\mathalpha{=}2$) & 96.66 & 58.77 & 71.96 & \textbf{68.13} & \textbf{73.88} \\
     Ours ($\hat{K}\mathalpha{=}3$) & \textbf{97.02} & 58.37 & 71.40 & 67.89 & 73.67\\
     Ours ($\hat{K}\mathalpha{=}4$) & 96.57 & 58.66 & \textbf{72.09} & 66.79 & 73.53 \\ \hline
    \end{tabular}
    }
    \caption{Results in the VLCS dataset. The respective scores are obtained from each method's original paper. For details about the meaning of columns and use of asterisks, see Table~\ref{tab:PACS}. 
    }
    \label{tab:VLCS}
\end{table}

\subsection{Results}
Table~\ref{tab:PACS} and Table~\ref{tab:VLCS} show the experimental results with the PACS and VLCS datasets, respectively. The scores shown in the tables are the average over five repetitions for each run, and $\hat{K}$ denotes the number of pseudo domains used in our method. For all datasets, our method achieves results that surpass those of existing methods regardless of the number of pseudo domains $\hat{K}$.
Below, we discuss the influence of the number of pseudo domains $\hat{K}$. In the PACS dataset, our method has a significant advantage with respect to the corresponding Deep All baseline. The results show that training the domain-invariant feature extractor using adversarial learning is effective for domain generalization among more diverse domains such as the PACS dataset. This good performance is achieved without using any domain labels, unlike other methods excluding JiGen. Our method can discover latent domains and assign pseudo domain labels by focusing on the image styles.\par
Moreover, even in the VLCS dataset, where domain shifts are inside only photo images, our method can improve the classification accuracy compared to other methods.
The results show that even if the original domain labels of datasets are not separated by the image styles, our method can improve the generalization performance by assigning pseudo domain labels by focusing on them.

\section{Further Analysis}
\begin{table}[t]
    \centering
    \resizebox{\columnwidth}{!}{
    \begin{tabular}{ccccc|c} \hline
     PACS & Art. & Cartoon & Sketch & Photo & Avg.\\ \hline
     \multicolumn{6}{c}{AlexNet}\\ \hline
     Deep All & 68.09 & 70.23 & 61.80 & 88.86 & 72.25\\
     Ours w/o $L_{\mathrm{adv}}$ & 67.66 & 70.45 & 62.56 & 88.94 & 72.40\\
     Ours w/o $L_{\mathrm{ent}}$ & 68.31 & 71.13 & 65.26 & \textbf{89.38} & 73.52\\
     Ours w/o stat. & 67.37 & 70.22 & 63.12 & 89.20 & 72.48 \\
     Ours w/o iter. & 69.13 & 70.72 & 65.41 & 89.11 & 73.59 \\
     Ours w/o clus. & 68.49 & 72.24 & 66.31 & 89.27 & 74.08 \\
     Ours & \textbf{69.27}& \textbf{72.83} & \textbf{66.44} & 88.98 & \textbf{74.38}\\ \hline
    \end{tabular}
    }
    \caption{Results of the ablation study in the PACS dataset. For details about the meaning of columns, see Table~\ref{tab:PACS}.}
    \label{tab:ablation}
\end{table}

\begin{figure*}
\hfill
\begin{minipage}{0.33\hsize}
\begin{center}
\includegraphics[width=1.0\linewidth]{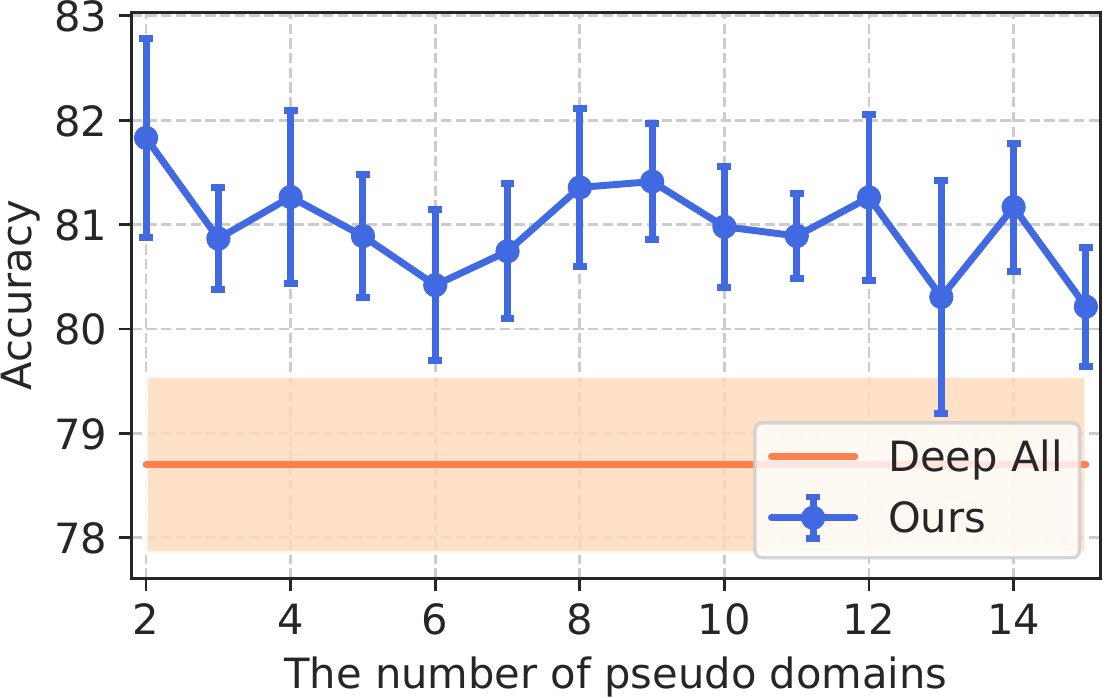}
\caption{Results obtained when varying the number of pseudo domains.
The accuracy is the average of five sets.}
\label{fig:cluster}
\end{center}
\end{minipage}
\hfill%
\begin{minipage}{0.33\hsize}
\begin{center}
\includegraphics[width=1.0\linewidth]{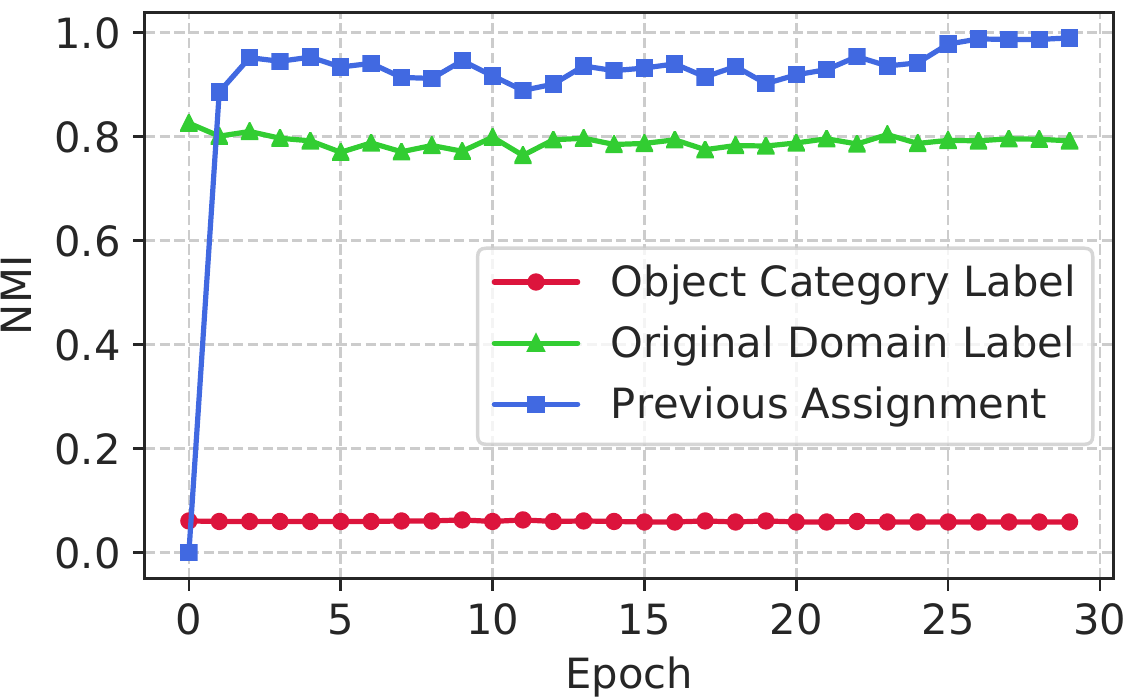}
\caption{NMI between pseudo domain labels and object category labels, original domain labels, and previous assignments.}
\label{fig:nmi}
\end{center}
\end{minipage}
\hfill%
\begin{minipage}{0.19\hsize}
\begin{center}
\includegraphics[width=1.0\linewidth]{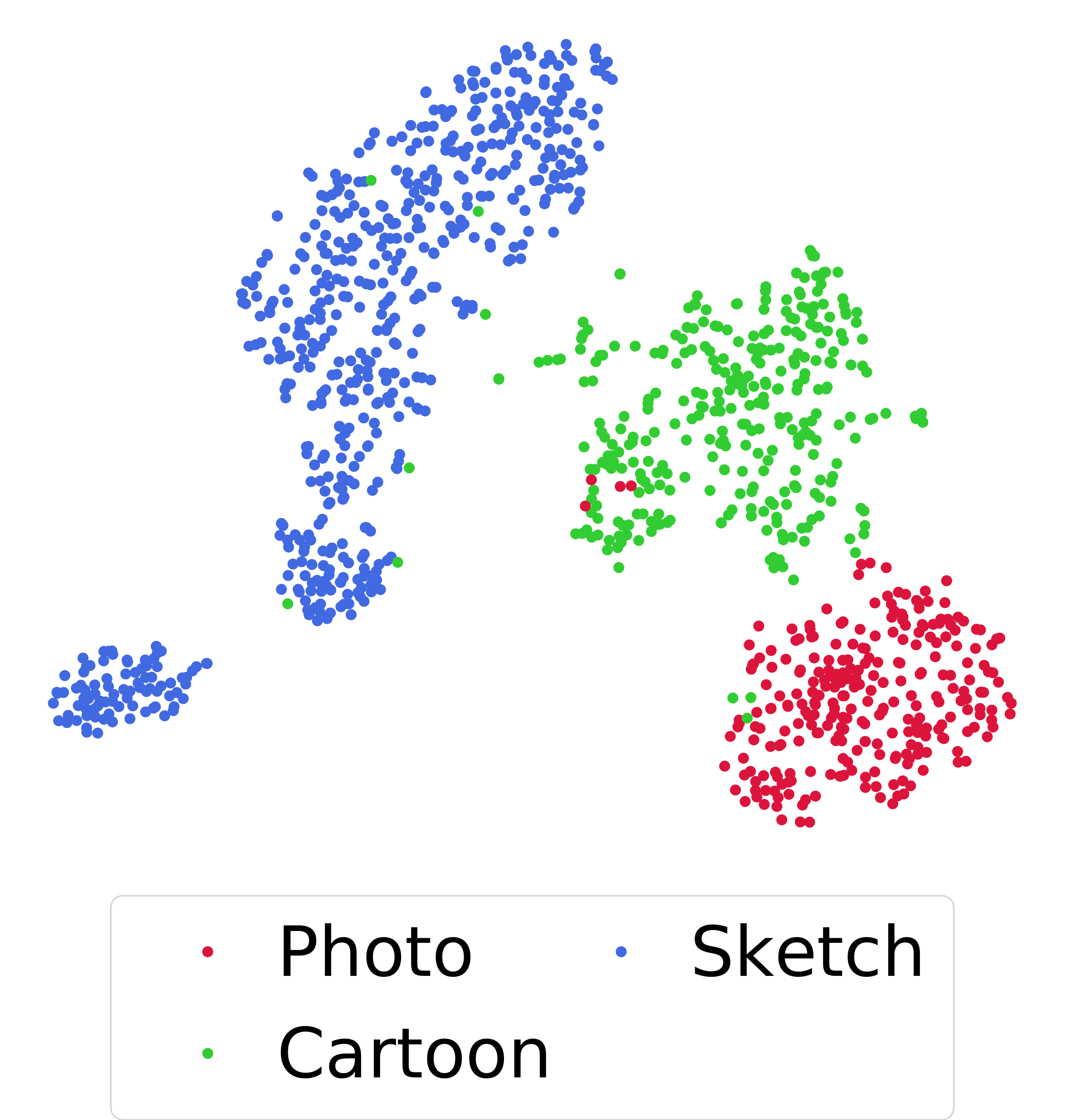}
\caption{T-SNE visualization of domain discriminative features.}
\label{fig:visualize}
\end{center}
\end{minipage}
\hfill\null
\end{figure*}

\subsection{Ablation Study}
In this section, we describe an ablation study to investigate the effect of different components of our method using the PACS dataset and AlexNet.
The variants of our method used in the experiments are as follows. Our method without $L_{\mathrm{adv}}$: The model that removes the adversarial loss in Eq.~\ref{eq:adv}. Our method without $L_{\mathrm{ent}}$: The model that removes the entropy loss in Eq.~\ref{eq:entropy}. Our method without stat.: The model that simply uses outputs of the convolutional layer (\texttt{relu2} in this experiment) as domain-discriminative features for clustering instead of a stack of convolutional feature statistics in Eq~\ref{eq:feature}. Our method without iter.: The model that uses the first assigned pseudo domain labels to the end without iteratively reassigning them. Our method without clus.: The model that uses original domain labels instead of assigning pseudo domain labels by clustering.\par 
Table~\ref{tab:ablation} shows the experimental results obtained when the number of pseudo domains is set to three. The results of our method without $L_{\mathrm{adv}}$ and our method without $L_{\mathrm{ent}}$ indicate that the adversarial loss in Eq.~\ref{eq:adv} is effective for domain generalization, and it is further improved by using the entropy loss in Eq.~\ref{eq:entropy}. The result of our method without stat. indicates that simply using the outputs of convolutional layers cannot sufficiently extract domain-discriminative features, and it cannot achieve domain generalization so well.
The result of our method without iter. indicates that iteratively reassigning pseudo domain labels improve the classification performance compared with those assigned at the start of training to the end. This may be because domain-discriminative features can be extracted more successfully by using models trained with samples of each domain, rather than using a pre-trained model.
Finally, the result of our method without clus. indicates that the use of iteratively reassigned pseudo domain labels improves the classification accuracy compared with the use of original domain labels. It appears that pseudo domain labels are suitable for training the domain-invariant feature extractor because they are based on the model's inner features and capture image styles.

\subsection{Varying the Number of Pseudo Domains}
In domain generalization using a mixture of multiple latent domains, the number of multiple latent domains in the source domain is unknown. 
Although our method divides samples into $\hat{K}$ pseudo domains by clustering, we have to set the number of pseudo domains in advance. It is unclear whether our method works accurately if the number of pseudo domains is not the same as the number of original domains.
Therefore, we check the performance of our method when changing the number of pseudo domains. We use the same experimental setting of the previous paragraph with the PACS dataset and ResNet-18. We consider four experiments in which the target domains are changed as one set, and repeat it five times.
Fig.~\ref{fig:cluster} shows the mean and standard deviation results of our proposed method and Deep All. Note that in reality, the number of original domains is three. Based on the results obtained, there is no significant correlation between the number of pseudo domains and the classification accuracy, which highlights the robustness of our method to varying numbers of pseudo domains.

\subsection{Clustering Evaluation}
Our method assigns pseudo domain labels by clustering. There is a concern that clustering is not performed by domains but by object categories, although it does not necessarily have to divide samples by original domains.
Therefore, we evaluate the clustering by calculating the normalized mutual information (NMI) between pseudo domain labels and object category labels, original domain labels, and pseudo domain labels before one epoch. Moreover, we visualize the distribution of domain-discriminative features using t-SNE \cite{t_SNE}. We use the same experiment setting of the previous paragraph with the PACS dataset and AlexNet, set the number of pseudo domains to three, and set Art-painting to the target domain.\par
Fig.~\ref{fig:nmi} shows that the NMI between pseudo domain labels and the original domain labels is large, while that between pseudo domain labels and object category labels is small. Moreover, the NMI between pseudo domain labels and original domain labels remains almost unchanged over the whole training period. These indicate that clustering domain-discriminative features divides samples not by object categories but original domains over the whole training period. This fact can also be seen in Fig.~\ref{fig:visualize}, where the distributions of domain-discriminative features are roughly divided by their original domains. Moreover, the NMI between pseudo domain labels and the previous assignment gradually converges to 1.0 as the training proceeds, which indicates that clustering results become gradually stable.

\section{Conclusion}
In this study, we proposed a new scenario called domain generalization using a mixture of multiple latent domains. To address this scenario, we proposed a new method that extracts a stack of convolutional feature statistics representing the image styles as domain-discriminative features, assigns pseudo domain labels by clustering them, and trains the domain-invariant feature extractor from among latent domains using adversarial learning. In the experiments, our method without domain labels achieved a better performance than conventional methods that use them.

\section{Acknowledgments}
This work was partially supported by JST CREST Grant Number JPMJCR1403, and partially supported by JSPS KAKENHI Grant Number JP19H01115. We would like to thank Yusuke Mukuta, Antonio Tejero de Pablos, Atsuhiro Noguchi, Akihiro Nakamura for helpful discussions.

\bibliographystyle{aaai}
\bibliography{bibfile}

\end{document}